\documentclass{article}

\PassOptionsToPackage{numbers, compress}{natbib}
\usepackage[final]{nips_2018}

\usepackage[utf8]{inputenc} % allow utf-8 input
\usepackage[T1]{fontenc}    % use 8-bit T1 fonts
\usepackage{hyperref}       % hyperlinks
\usepackage{url}            % simple URL typesetting
\usepackage{booktabs}       % professional-quality tables
\usepackage{amsfonts}       % blackboard math symbols
\usepackage{nicefrac}       % compact symbols for 1/2, etc.
\usepackage{microtype}      % microtypography
\usepackage{todonotes}
\usepackage{amsmath, amssymb}
\usepackage{subcaption}
\usepackage{enumitem}
\DeclareMathOperator*{\argmax}{arg\,max}

\newcommand{\bg}{\boldsymbol{g}}

\newcommand{\bp}{\boldsymbol{p}}

\newcommand{\br}{\boldsymbol{r}}
\newcommand{\bs}{\boldsymbol{s}}

\newcommand{\bx}{\boldsymbol{x}}

\newcommand{\bz}{\boldsymbol{z}}

\newcommand{\bE}{\boldsymbol{E}}
\newcommand{\bF}{\boldsymbol{F}}

\newcommand{\bL}{\boldsymbol{L}}

\newcommand{\bO}{\boldsymbol{O}}

\newcommand{\bR}{\boldsymbol{R}}

\title{A Probabilistic Model of Cardiac \\ Physiology and Electrocardiograms}

\author{
  Andrew~C.~Miller\thanks{\texttt{am5171@columbia.edu}, \url{http://andymiller.github.io/}} \\
  Columbia University \\
  \And
  Ziad Obermeyer \\
  University of California at Berkeley \\
  \And
  David~M.~Blei \\
  Columbia University \\
  \And 
  John~P.~Cunningham \\
  Columbia University \\
  \And
  Sendhil Mullainathan \\
  University of Chicago \\
}

\begin{document}

\maketitle

\begin{abstract}
An electrocardiogram (EKG) is a common, non-invasive test that measures the electrical activity of a patient's heart.
EKGs contain useful diagnostic information about patient health that may be absent from other electronic health record (EHR) data.
As multi-dimensional waveforms, they could be modeled using generic machine learning tools, such as a linear factor model or a variational autoencoder. 
We take a different approach:~we specify a model that directly represents the underlying electrophysiology of the heart and the EKG measurement process. 
We apply our model to two datasets, including a sample of emergency department EKG reports with missing data.
We show that our model can more accurately reconstruct missing data (measured by test reconstruction error) than a standard baseline when there is significant missing data.
More broadly, this physiological representation of heart function may be useful in a variety of settings, including prediction, causal analysis, and discovery.
\end{abstract}

%-------------------------------------------------%
% Intro                                           %
%-------------------------------------------------%
\section{Introduction}
Heart disease annually claims the lives of over 600{,}000 people in the United States and over 17 million people worldwide \citep{cdc2017heart, who2017heart}.
Early detection of cardiovascular disease is critical, and accurate characterization of patient risk can improve care.
Consequently, a commonly used diagnostic tool is the electrocardiogram (EKG), which uses electrodes placed on the patient's body to measure the heart's electrical activity. 
EKGs are widely administered to diagnose a variety of cardiac abnormalities, including ischemia, heart attack, and arrhythmias.  
EKGs (and other waveform and image data) likely carry additional information about the overall health of a patient, beyond what is already in electronic health records (e.g.~past diagnoses, medications, and lab tests). 
As we begin to use EHR data to inform health care decisions, EKG features can be used alongside patient EHR data to construct more accurate predictors, similar to unstructured physician notes \citep{rumshisky2016predicting} or the dynamics of ICU vital signs \citep{ghassemi2017predicting}. 

However, unlike discrete records, the EKG is a high-dimensional, dynamic object that is difficult to summarize. 
While an EKG could be summarized with standard ML tools, a generic approach would ignore the underlying physiology that drives EKG observations: the waveforms are a measurement of an electrical signal originating in the heart and propagating through the body. 

In this work, we pursue a physiological representation of an EKG by modeling the location, orientation, and strength of the electric potential induced by cardiac activity. 
Such a model has many benefits. 
First, because our model represents physical electrode locations, our approach can flexibly and coherently incorporate data from systems with more (or fewer) electrodes than standard 12-lead EKGs, including higher-dimensional (e.g.256-electrode ~body surface EKGs \citep{cluitmans2015noninvasive}) and lower-dimensional (e.g.~5-electrode Holter monitors) electrocardiographic signals.  
Second, we show that our model is robust to large amounts of missing observations.
Finally, it produces a physically interpretable representation of the EKG that can be used in a variety of applications, such as prediction or causal analysis --- for example, measuring the effect of a therapy on some aspect of cardiac function.

We evaluate our model with a lead reconstruction task --- given an EKG where data from some leads are missing, how accurately can we reconstruct the held out data? 
We find that our model can more accurately reconstruct missing EKG leads than non-physical factor models when we have a lot of missing data (which is common in our clinical emergency department setting).
We conclude with a discussion of statistical inference challenges and model extensions.

%-------------------------------------------------%
% Problem setup/formalization                     %
%-------------------------------------------------%
\section{Data Description and Model}
\label{sec:ekg-model}
\paragraph{EKG Background} An EKG lead measures the electric potential difference between two (or more) electrodes. 
An \emph{electrode} is a conductive pad placed on a part of the body (e.g.~left leg or chest) that measures instantaneous electric potential.
A \emph{lead} records the difference between a pair (or a specific combination) of electrode readings (typically in millivolts); cardiologists visually inspect the lead outputs for abnormalities.
The standard twelve leads are commonly referred to as ${\bL = \{I, II, III, aVR, aVL, aVF, V1, \dots, V6 \}}$.\footnote{There are typically nine electrodes used to produce a twelve lead EKG --- see Appendix~\ref{sec:electrode-to-lead-transformation}.}
An EKG tracing of length $T$ consists of $|\bL|=12$-dimensional observations at $T$ time steps,
\begin{align}
    \bx &= (\bx_{1}, \dots, \bx_{T}) \, \quad \bx_{t} \in \mathbb{R}^{|\bL|} \, .
\end{align}
We model lead EKG observations by describing the physical location, orientation, and strength of the electric potential induced by cardiac activity.
We introduce model components that (i) describe the underlying state of electrical activity; (ii) map latent electrical activity to electrode potential values; and (iii) map electrode values to lead observations.
To address point (i) we model electrical activity with a single moving dipole, following \citep{okamoto1982moving}.
We address point (ii) by modeling the 3-d location of electrodes and using a simplified torso conductance model.
Point (iii) is straightforward and can be modeled by a fixed linear map --- we describe this step in detail in Appendix~\ref{sec:electrode-to-lead-transformation}.

%---------------------------------%
% Observation Model FIgure        %
%---------------------------------%
\begin{figure}[t!]
    \centering
    \begin{subfigure}[b]{0.43\linewidth}
        \centering
        \includegraphics[width=.61\textwidth]{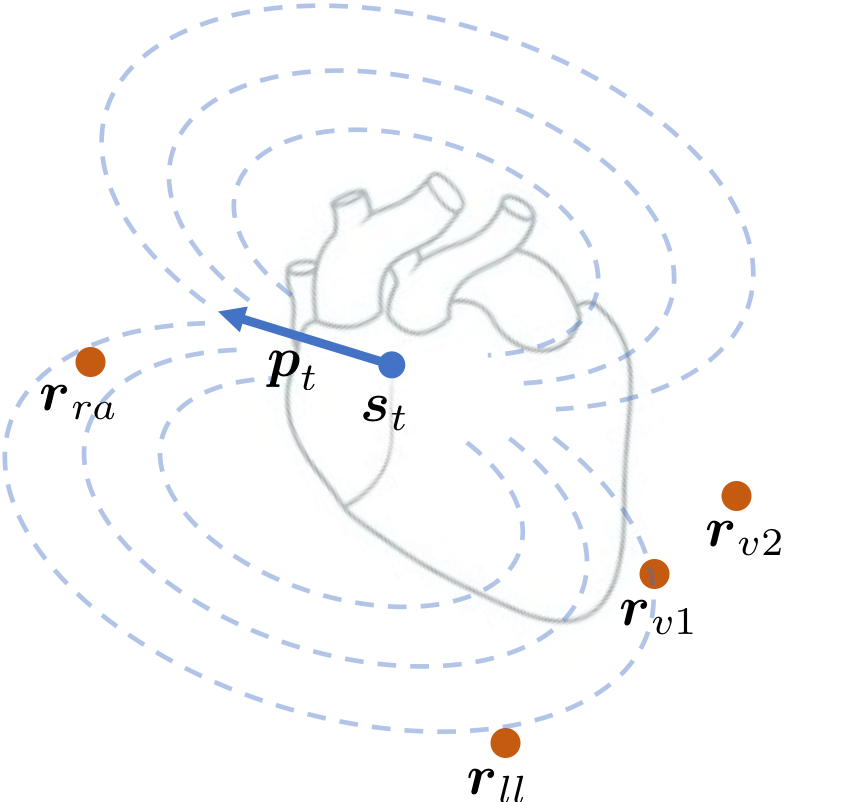}
        \caption{}
        \label{fig:dipole-cartoon}
    \end{subfigure}%
    \begin{subfigure}[b]{0.43\linewidth}
        \centering
        \includegraphics[width=\textwidth]{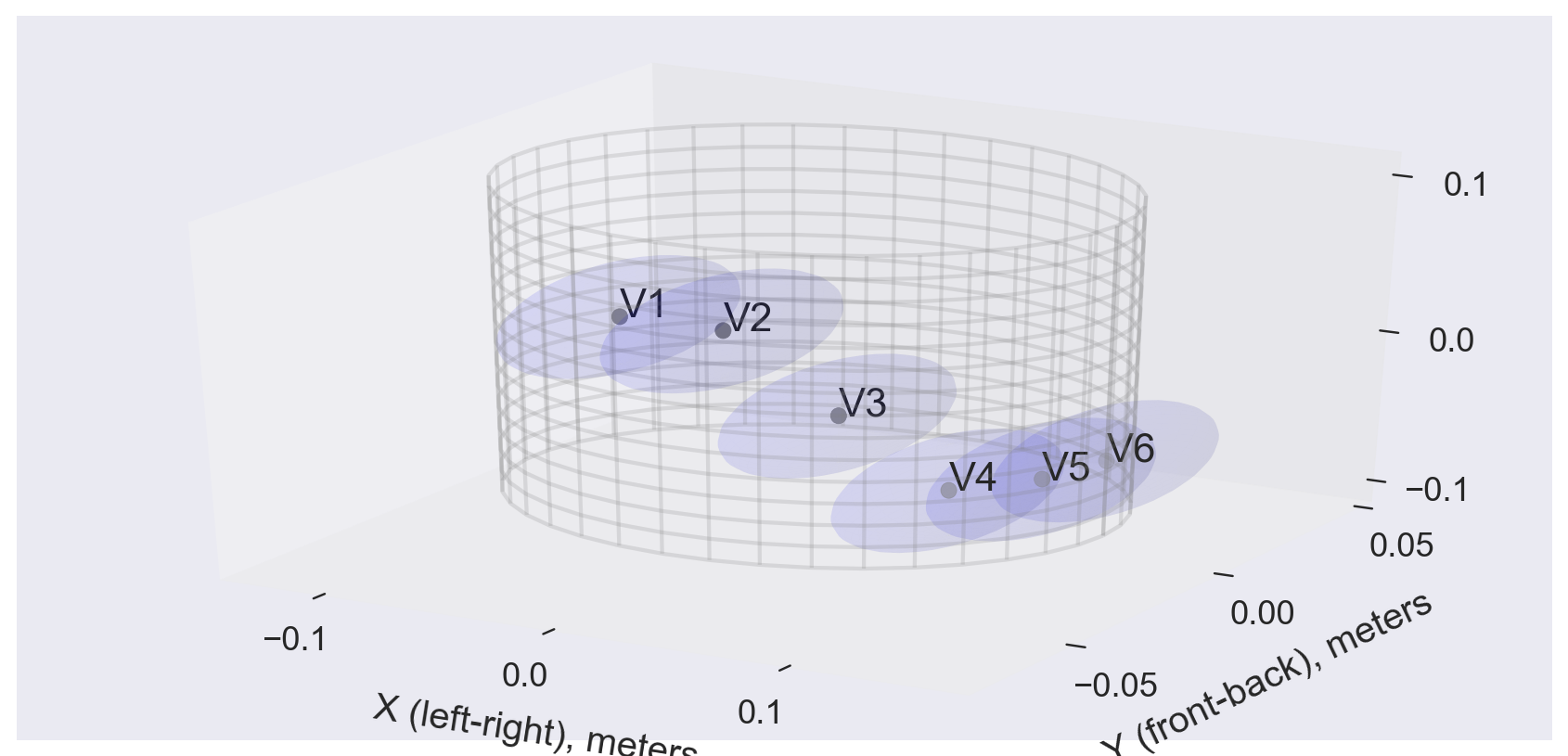}
        \caption{}
        \label{fig:precordial-electrode-prior}
    \end{subfigure}%
    \caption{(a) Graphical depiction of the latent moving dipole model.  The instantaneous electrical activity of the heart is summarized by a latent moving dipole, with location $\bs_t$ and moment $\bp_t$.  This object induces a potential field (depicted by dotted lines), which is measured by the electrodes located at $\br_{e}$ according to Equation~\ref{eq:electrode} --- we depict the right arm ($e=ra$), left leg ($e=ll$) and two precordial electrodes, $e=v1$ and $e=v2$. 
    The electrode measurements are linearly combined to produce the EKG lead observations (mentioned in Section~\ref{sec:ekg-model} and detailed in Appendix~\ref{sec:electrode-to-lead-transformation}). 
    (b) Prior distribution over the 3-d spatial location of the precordial electrodes, $v1$ - $v6$. 
    }
    \label{fig:model-description}
    \vspace{-.35em}
\end{figure}

%----------------------------
% Model Description 
%----------------------------
\vspace{-.75em}
\paragraph{Moving Dipole Generative Model}
Each observation $\bx_t$ is generated by the electric potential induced by a single latent dipole with state $\bz_t \triangleq (\bs_t, \bp_t)$, where $\bs_t \in \mathbb{R}^3$ is the spatial location of the dipole and $\bp_t \in \mathbb{R}^3$ is its moment.
A dipole is an idealized point whose location and moment induce a spatial electrostatic potential that is measured by each electrode. 
At time $t$, the latent dipole induces a potential at each of the $|\bE|$ electrodes that are spatially located at points $\br_e$ where $e=1,\dots,|E|$ and $\br_e \in \mathbb{R}^3$.
Assuming a uniform conductance torso model, a single dipole with state $\bz_t = (\bs_t, \bp_t)$ will have a potential measured at electrode $e$, located at $\br_{e}$, given by 
\begin{align}
    \tilde{x}_{t,e} &= \frac{1}{4\pi\kappa} \cdot \frac{(\br_{e}-\bs_t)^\intercal \bp_t }{||\br_{e}-\bs_t||^3} \triangleq g(\bs_t, \bp_t, \br_e, \kappa) \, ,
    \label{eq:electrode}
\end{align}
where $\kappa$ is the (constant) electrical conductivity of the torso \citep{okamoto1982moving} (which we set to $.2$, following \citep{sovilj2013simplified}).
To get from electrode to lead, there is a fixed linear function which produces the twelve standard leads, namely $\bx_t = \bO \tilde{\bx}$, where $\tilde{\bx}_t = (\tilde{x}_{t,1}, \dots \tilde{x}_{t,|E|})$ and $\bO$ is detailed in Appendix~\ref{sec:electrode-to-lead-transformation}.

\vspace{-.75em}
\paragraph{Priors}
To complete the model, we specify a prior distribution over the latent dipole state, $\bs_t$ and $\bp_t$, and the location of each electrode $\br_e$.
We put zero-centered spherical Gaussian priors over $\bs_t$ and $\bp_t$ and weakly informative priors over electrode locations, $p_e(\br)$, described in Appendix~\ref{sec:location-prior}.
The data generating process is 
\begin{align}
    \br_{e} &\sim p_e(\br) &&\text{ for } e = 1, \dots, |E| \\
    \bs_t, \bp_t &\sim \mathcal{N}(0, I\sigma_{\bs}^2), \mathcal{N}(0, I\sigma_{\bp}^2) &&\text{ for } t = 1, \dots, T\\
    \bx_t &\sim \mathcal{N}(\bO \tilde{\bx}_t, I\sigma^2) &&\text{ where } \tilde{\bx}_{t,e} = g(\bs_t, \bp_t, \br_e, \kappa)
    \label{eq:likelihood}
\end{align}
In this current iteration we ignore temporal correlation in $\bs_t$ and $\bp_t$, leaving this for future work.

\vspace{-.75em}
\paragraph{Inference}
The above generative model will admit tractable (though approximate) variational inference \citep{ranganath2014black}.
As a first step, however, we use MAP inference to show the utility of this modeling approach, solving the optimization problem
\begin{align}
    \{ \hat \bz \}, \{\hat \br\} &= \argmax_{\bz, \br} \ln p(\bx, \bz | \br) p(\br) \\
    \ln p(\bx, \bz | \br) p(\br) &= \ln \sum_{t=1}^T \ln p(\bx_t | \bz_{t}, \br) + \ln p(\bz_t) + \sum_{e=1}^{|E|} \ln p_e(\br_e) \, .
\end{align}
To solve this optimization problem, we use L-BFGS \citep{liu1989limited} with gradients computed by automatic differentiation \citep{maclaurin2015autograd}. 
See Section~\ref{app:inference} for further discussion of inference.

%----------------------------------------------------%
% experiment set up figure --- missing ness scheme   %
%----------------------------------------------------%
\begin{figure}[t!]
    \centering
    \begin{subfigure}[b]{0.49\linewidth}
        \centering
        \includegraphics[width=\textwidth]{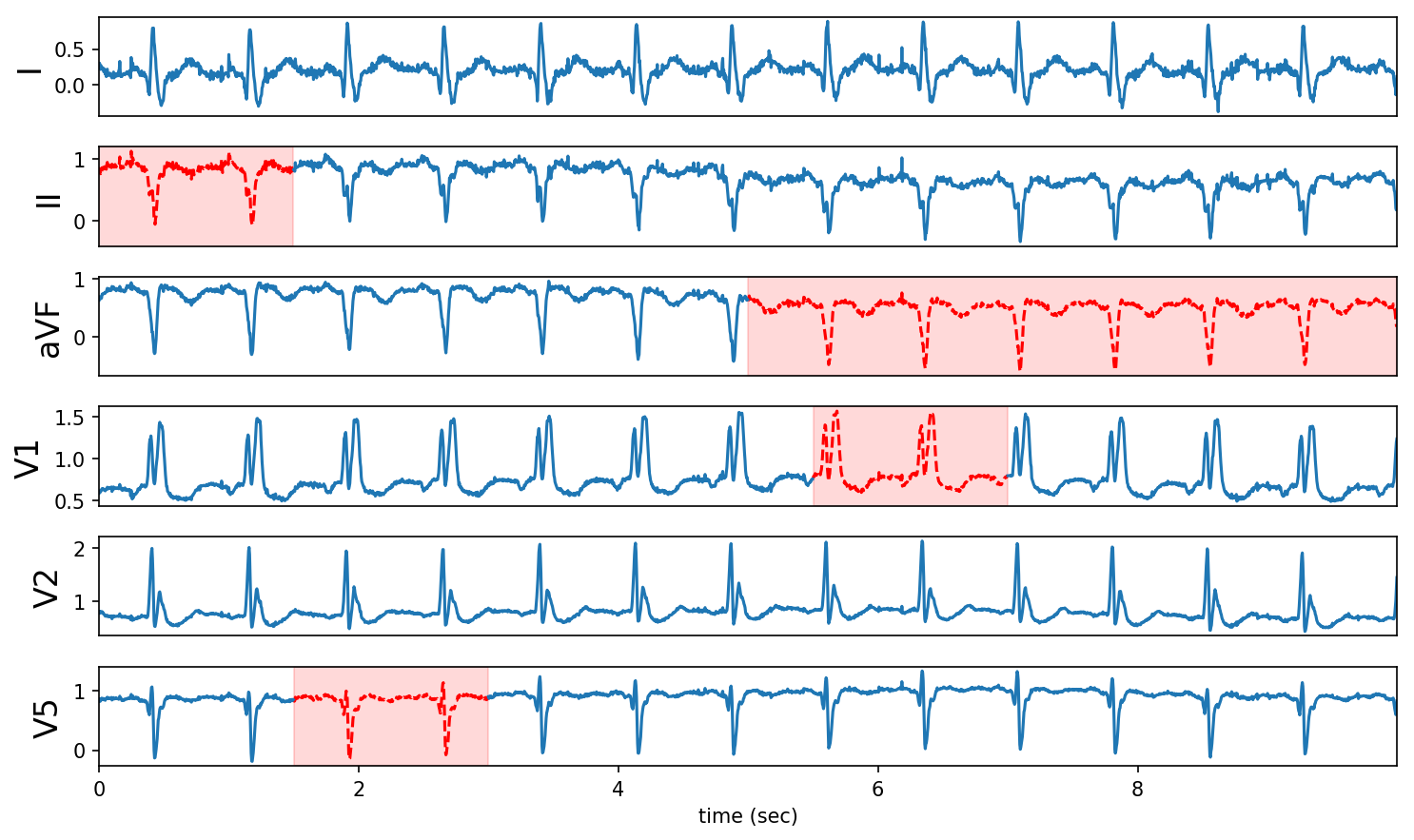}
        \caption{\texttt{ptb} example}
        \label{fig:ptb-example}
    \end{subfigure}%
    \begin{subfigure}[b]{0.49\linewidth}
        \centering
        \includegraphics[width=\textwidth]{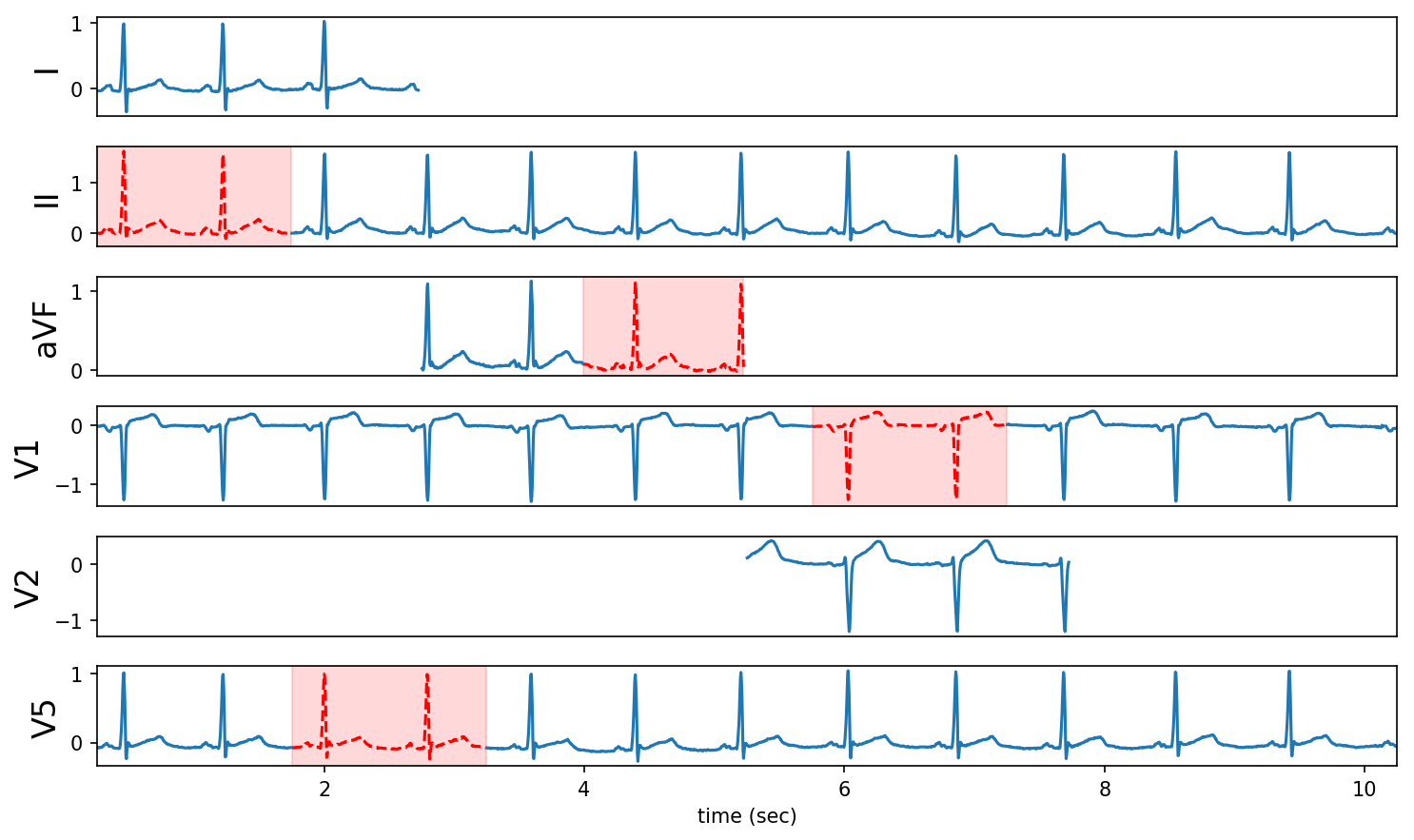}
        \caption{\texttt{ed} example}
        \label{fig:ed-example}
    \end{subfigure}%
    \caption{Observed and held out data. We fit each model on the observed (solid blue) signal and test on the held out (dashed red) areas.  \emph{Left}: \texttt{ptb} examples have data from all twelve leads over the observation period. \emph{Right}: \texttt{ed} examples have three ``long leads'' (II, V1, V5) and observe roughly 2.5 second chunks of every other lead, which yields large amounts of missing data.
    Note that only six of the twelve leads are depicted above.}
    \label{fig:holdout-scheme}
    \vspace{-.35em}
\end{figure}

%-------------------------------------------------%
% Experiments                                     %
%-------------------------------------------------%
\section{Experiments}

We assess model performance by measuring held-out lead reconstruction error. 
For each EKG, we hold out portions of the waveform during training (depicted in Figure~\ref{fig:holdout-scheme}).
We then use the model to impute the held-out values and compute the root mean squared error (RMSE).
A model that better captures the underlying phenomenon will more accurately impute missing lead values.

We apply our model to a small sample of 10-second EKG segments from two datasets: 
\vspace{-.6em}
\begin{itemize}[leftmargin=*] \itemsep 1pt
    \item \texttt{ptb}:  the PTB Diagnostic ECG Database \citep{goldberger2000physiobank}, a database of 549 full twelve-lead EKG records.  Figure~\ref{fig:ptb-example} displays a typical \texttt{ptb} record. 
    \item \texttt{ed}: a subset of 997 EKG recordings from emergency department patients from a large metropolitan hospital, each exhibiting ``normal sinus rhythm.''  These EKGs are derived from a traditional report that saves the full time series for leads II, V1, and V5, and short (approximately 2.5 second) segments of the other 9 leads.  This missingness can frustrate analysis of the full EKG object, which we address with both our model and baseline approaches.  Figure~\ref{fig:ed-example} shows a typical \texttt{ed} record. 
\end{itemize}

\vspace{-.75em}
\paragraph{Baseline linear factor model}
We compare the latent dipole model to probabilistic PCA (with $K=3$ and $K=6$ latent dimensions), which describes lead observations as 
\begin{align}
    %\bz_t &\sim \mathcal{N}(0, I_K) \\
    \bx_t &= \bF \bz_t + \epsilon \, , \quad  \text{ where } \bF \in \mathbb{R}^{12 \times K}, \bz_t \in \mathbb{R}^K \, , \epsilon \sim \mathcal{N}\left(0, \sigma^2 I \right) \, .
\end{align}
Here $\bF$ are the $K$ latent factors, and $\bz_t$ is the $K$-dimensional latent state that describes observation $\bx_t$ (analogous to our dipole location and moment).
We use an implementation of PCA designed to handle missing entries \citep{bailey2012principal}. 

\begin{figure}[t!]
    \centering
    \begin{subfigure}[b]{.49\textwidth}
        \centering
        \includegraphics[width=.78\textwidth]{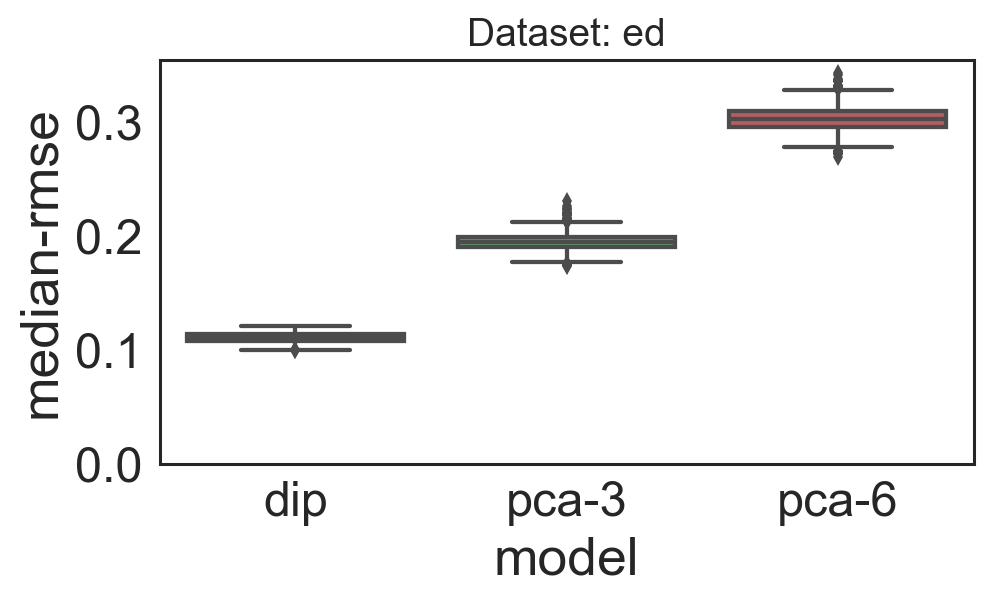}
        \caption{median rmse: \texttt{ed}}
        \label{fig:median-rmse-ed}
    \end{subfigure}%
    \begin{subfigure}[b]{.49\textwidth}
        \centering
        \includegraphics[width=.78\textwidth]{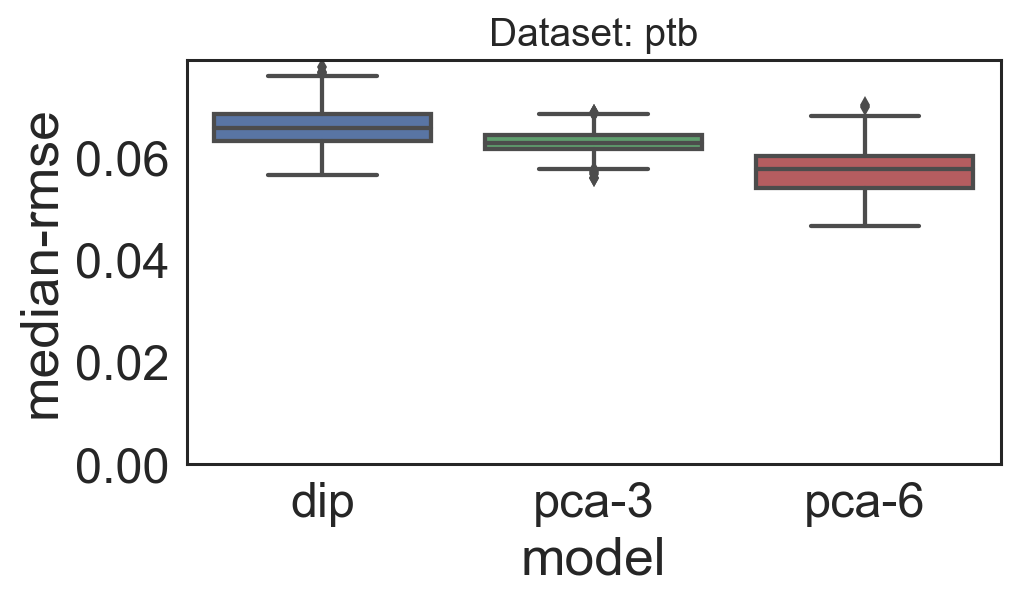}
        \caption{median rmse: \texttt{ptb}}
        \label{fig:median-rmse-ptb}
    \end{subfigure}  
    \caption{Median RMSE on held out leads for the dipole, PCA-3, and PCA-6 models (distributions computed with 1{,}000 bootstrap samples) on the \texttt{ed} (997 EKGs) and \texttt{ptb} (549 EKGs).
    The dipole model outperforms the PCA baselines in the \texttt{ed} dataset (which has a lot of missing data), while they perform similarly on the \texttt{ptb} dataset.
    }
    \label{fig:rmses}
    \vspace{-.5em}
\end{figure}

\vspace{-.75em}
\paragraph{Empirical Results}
For each dataset, we fit the dipole model and two PCA models ($K=3$ and $K=6$) to each EKG. 
We compare the median RMSE over all patients in each dataset.
Figure~\ref{fig:rmses} depicts the median RMSE for each model for the \texttt{ed} and \texttt{ptb}.
We observe that the dipole model more reliably reconstructs out of sample lead observations in the \texttt{ed} setting, where the short leads are missing most entries (see Figure~\ref{fig:holdout-scheme}).
We observe similar performance for all three models in the \texttt{ptb} dataset.
The dipole model is constrained to produce a spatially consistent reconstruction of EKG leads, which may help in the absence of many lead observations.

We produce a full imputation of an \texttt{ed} EKG in  Figure~\ref{fig:full-ed-reconstruction}, which shows that the dipole model can have more stable imputations.
To contrast this, the \texttt{ptb} EKG imputation in Figure~\ref{fig:full-ptb-reconstruction} shows that both models have an easier time reconstructing out of sample leads when most of the tracing is present. 
Figure~\ref{fig:test-recon-ed} shows the dipole model reconstruction on an \texttt{ed} EKG.
Figure~\ref{fig:test-residual-ed} compares the residuals for the dipole model and two PCA models on the same EKG. 
We see that the dipole model accurately reconstructs leads II and aVF (almost no structure in the residual), but has a difficult time with the precordial leads V1 and V5, which may be due to inaccurate reconstruction of electrode location. 
The PCA models have structured residuals for all leads around regions of high voltage.

%----------------------------%
% Hold out RMSE analysis     %
%----------------------------%

%-------------------------------------------------%
% Discussion                                      %
%-------------------------------------------------%
\section{Discussion}
We developed a model of EKG observations with a physiological inductive bias. 
We compared our model to non-physical latent factor models, showing its potential to more accurately characterize patient physiology.
There are many directions of future work.
We will model temporal dynamics in the latent variables $\bz_t$, incorporating local temporal smoothness and quasi-periodicity into the prior.
Such dynamics may better characterize arrhythmias or other abnormalities not present in normal sinus rhythm EKGs. 
The non-conjugacy of the dipole-electrode model complicates inference; a focus will be on efficient approximate inference methods for this model, following \citep{gao2016linear} and \citep{johnson2016composing}. 
We also aim to use model output in downstream statistical analyses --- exploring the structure of patients with adverse cardiac events, building risk predictors alongside historical patient medical record that can aid physician decision-making, and measuring the causal effect of therapies on aspects of cardiac function. 

%---- references -----
{\small
\bibliography{refs.bib}
}

%---- appendix -----
\clearpage
\appendix
\section{Electrode to Lead transformation}
\label{sec:electrode-to-lead-transformation}
Electrocardiogram data are presented as leads, which are differences in potential values between electrode measurements (or averages of electrode measurements).  
This is a somewhat straightforward transformation that we will account for in the likelihood. 
The output potential difference for each \emph{lead} are simple functions of the electrode readings: 
\begin{align*}
    I   &= la - ra & aVR = ra - \frac{1}{2}(la + ll)\\
    II  &= ll - ra & aVL = la - \frac{1}{2}(ra + ll)\\
    III &= ll - la & aVF = ll - \frac{1}{2}(ra + la) \\
    V_{i} &= v_{i} - \frac{1}{3}(ra + la + ll)
\end{align*}
We can express these relationships as a simple linear transformation of the electrode values $\tilde{\bx}$ into lead values $\bx$
\begin{align}
    \bx_t = \bO \tilde{\bx}_t \, , \quad
    \bO = 
    \begin{pmatrix}
      \bO_{e} &  0_{3 \times 6} \\
      \bO_{a} &  0_{3 \times 6} \\
      \bO_{v} &  I_{6 \times 6}  
    \end{pmatrix} \in \mathbb{R}^{12 \times 9}
\end{align}
where the transformation matrix $\bO$ has a fixed block structure, where the blocks are defined
\begin{align}
    \bO_{e} = 
    \begin{pmatrix}
      1    & -1   & 0  \\
      0    & -1   & 1  \\
      -1   & 0    & 1
    \end{pmatrix} \, , 
    \quad
    \bO_{a} = 
    \begin{pmatrix}
      -1/2 & 1    & -1/2 \\
      1    & -1/2 & -1/2 \\
      -1/2 & -1/2 & 1    
    \end{pmatrix} \, , 
    \quad
    \bO_{v} = 
    \begin{pmatrix}
      -1/3 & -1/3 & -1/3  \\
      \dots & \dots & \dots \\
      -1/3 & -1/3 & -1/3   
    \end{pmatrix} \, .
\end{align}

\section{Priors}
\subsection{Electrode location prior}
\label{sec:location-prior}
Fitting data with the above appearance model will require inferring the electrode locations $\bR$. 
We can use the general guidelines of where to attach electrodes to place a prior on their relative location.  

We use a simple elliptical model of the thorax to set prior values on the location of V1 - V6.  We set the chest width to be 25 centimeters and the major-minor axis ratio to be 2.75.
The V1 - V6 locations are at even angular locations from 260 degrees to 360 degrees (all the way to the left side of the chest). 
Figure~\ref{fig:precordial-electrode-prior} depicts the prior (with 2 standard deviations for the precordial (chest) electrodes (v1 - v6).

The location of the right arm, left arm, and left leg electrodes will be given diffuse priors flanking the modeled torso depicted in Figure~\ref{fig:precordial-electrode-prior}. 

\subsection{Conductance Parameter Priors}

Average thoracic tissue impedance can be approximated as resistive with conductivity values around .2 S/m \citep{geselowitz1989theory}. 

\section{Additional Model Fit}
\begin{figure}[t!]
    \centering
    \begin{subfigure}[b]{\linewidth}
        \centering
        \includegraphics[width=\textwidth]{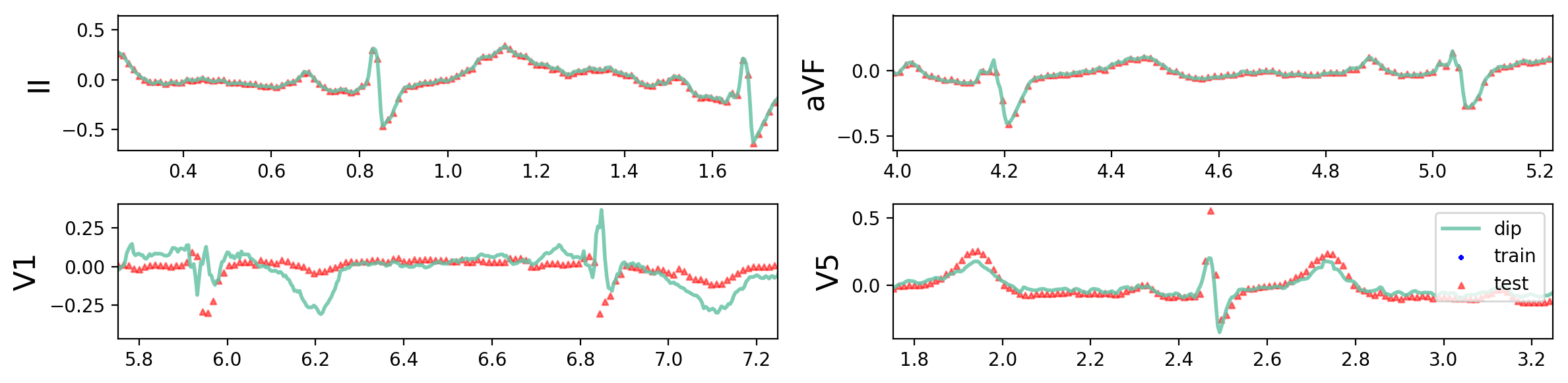}
        \caption{test reconstruction (\texttt{ed} example)}
        \label{fig:test-recon-ed}
    \end{subfigure}

    \begin{subfigure}[b]{\linewidth}
        \centering
        \includegraphics[width=\textwidth]{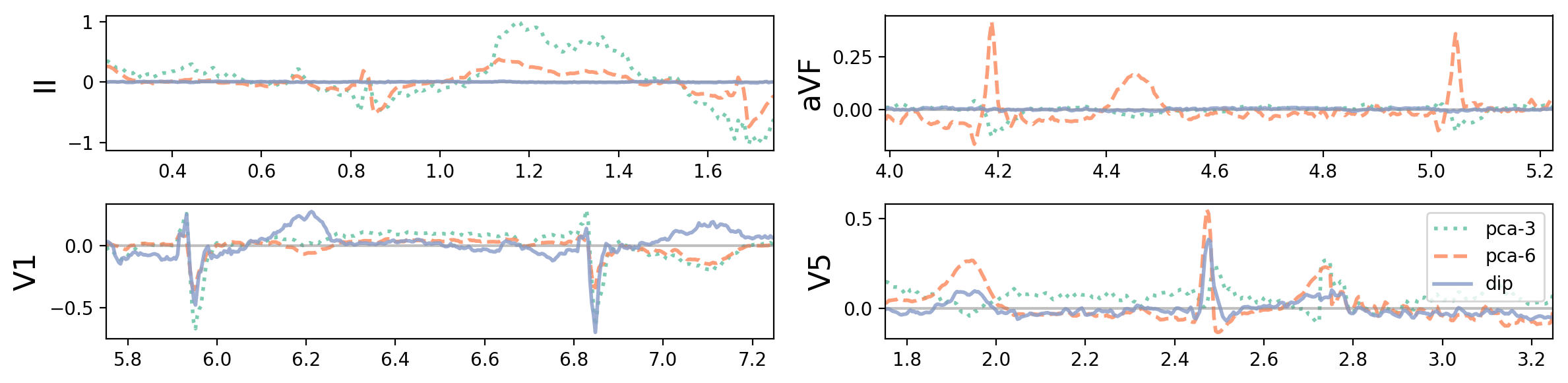}
        \caption{test residuals (\texttt{ed} example)}
        \label{fig:test-residual-ed}
    \end{subfigure}%
    \caption{(a) dipole model test reconstruction. (b) residual comparison for pca-3, pca-6, and dipole model.  }
    \label{fig:model-fit}
\end{figure}

\begin{figure}
    \centering
    \begin{subfigure}[b]{0.9\linewidth}
        \centering
        \includegraphics[width=\textwidth]{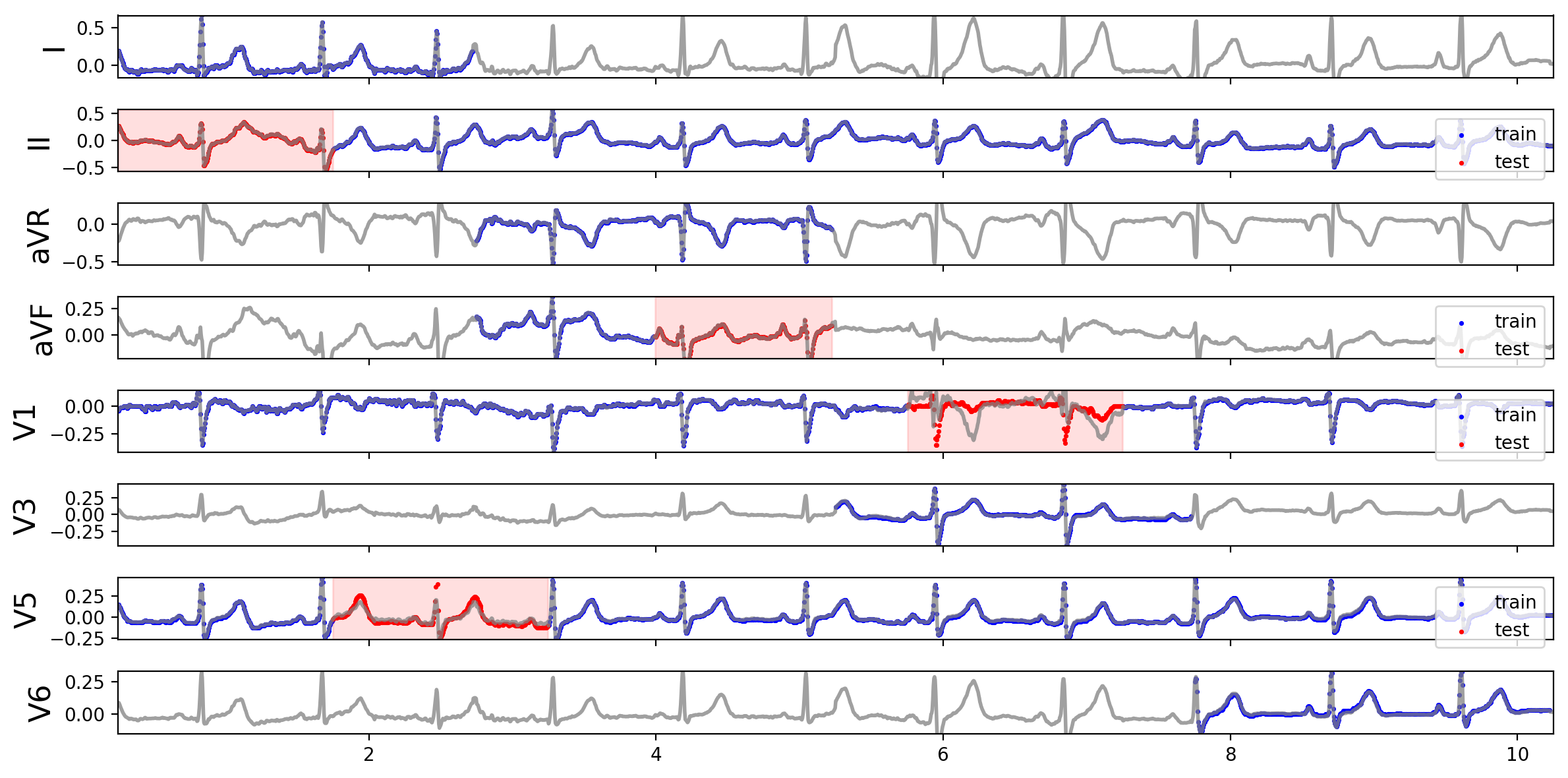}
        \caption{Dipole model}
    \end{subfigure}
    
    \begin{subfigure}[b]{0.9\linewidth}
        \centering
        \includegraphics[width=\textwidth]{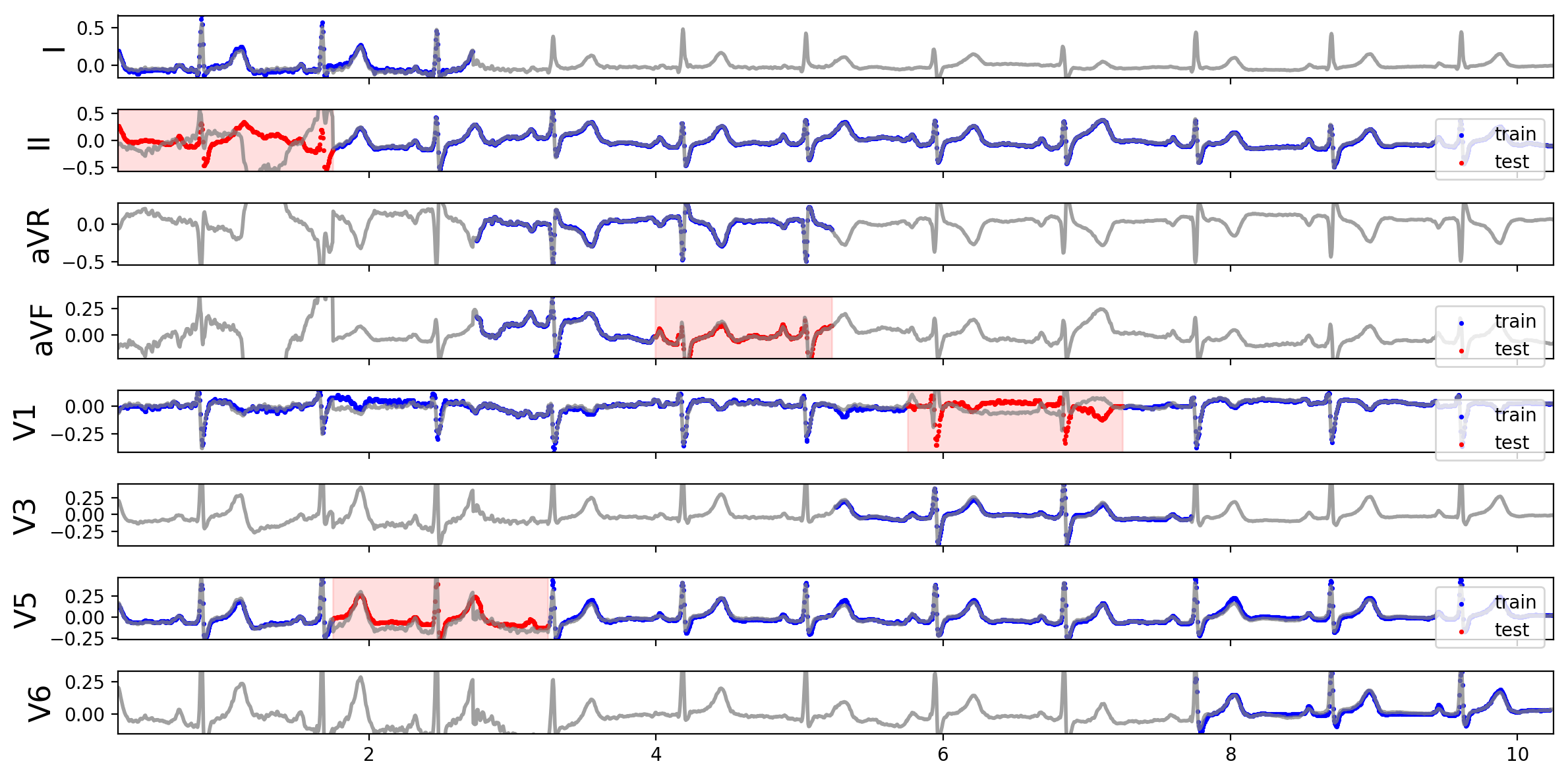}
        \caption{PCA $K=3$}
    \end{subfigure}%
        
    \begin{subfigure}[b]{0.9\linewidth}
        \centering
        \includegraphics[width=\textwidth]{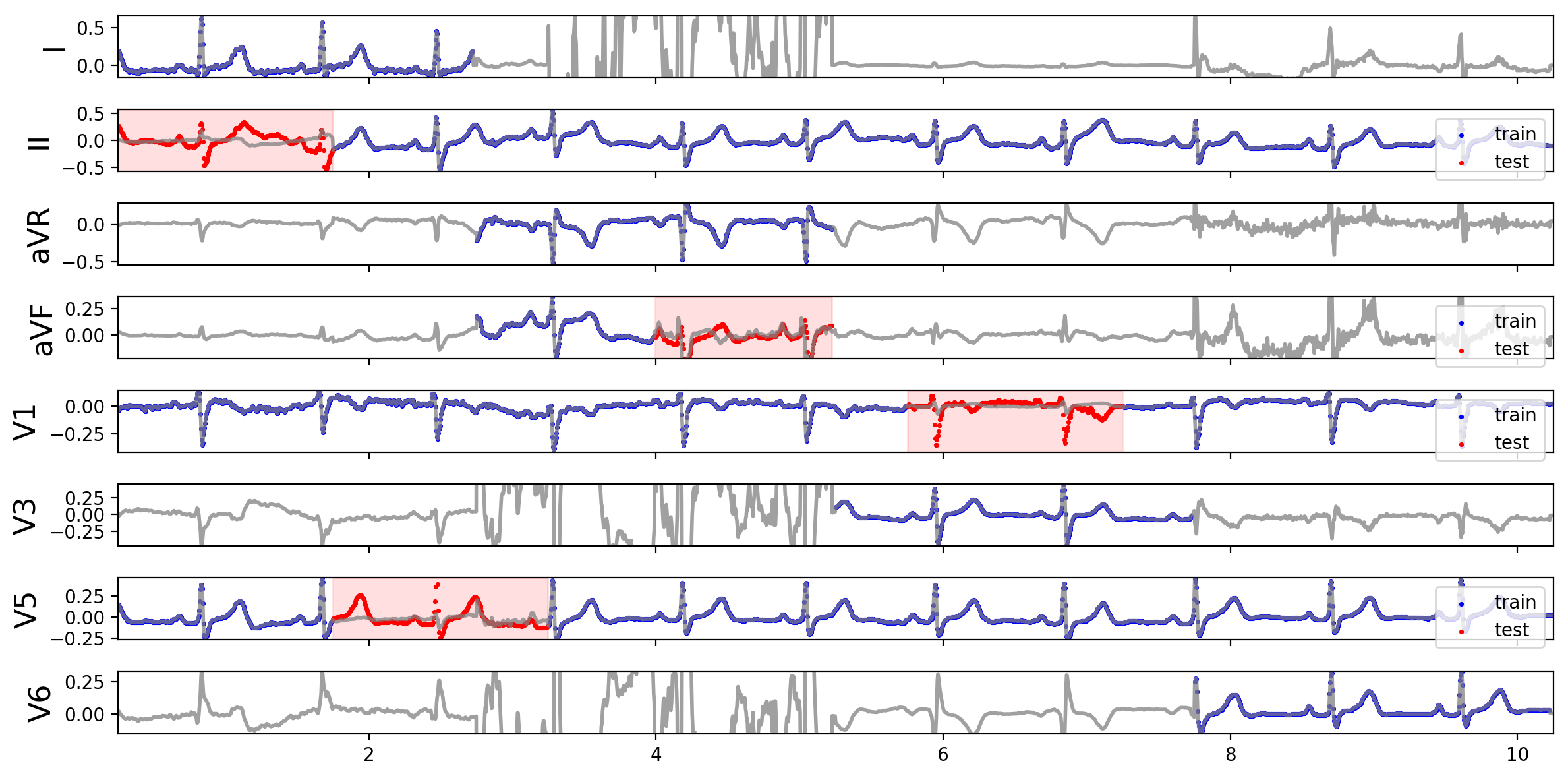}
        \caption{PCA $K=6$}
    \end{subfigure}%
    \caption{
    Held out and missing reconstruction for \texttt{ed} data (depicting only 8 leads). 
    Held out sections were selected to be non-overlapping. 
    }
    \label{fig:full-ed-reconstruction}
\end{figure}

\begin{figure}[t!]
    \centering
    \begin{subfigure}[b]{0.9\linewidth}
        \centering
        \includegraphics[width=\textwidth]{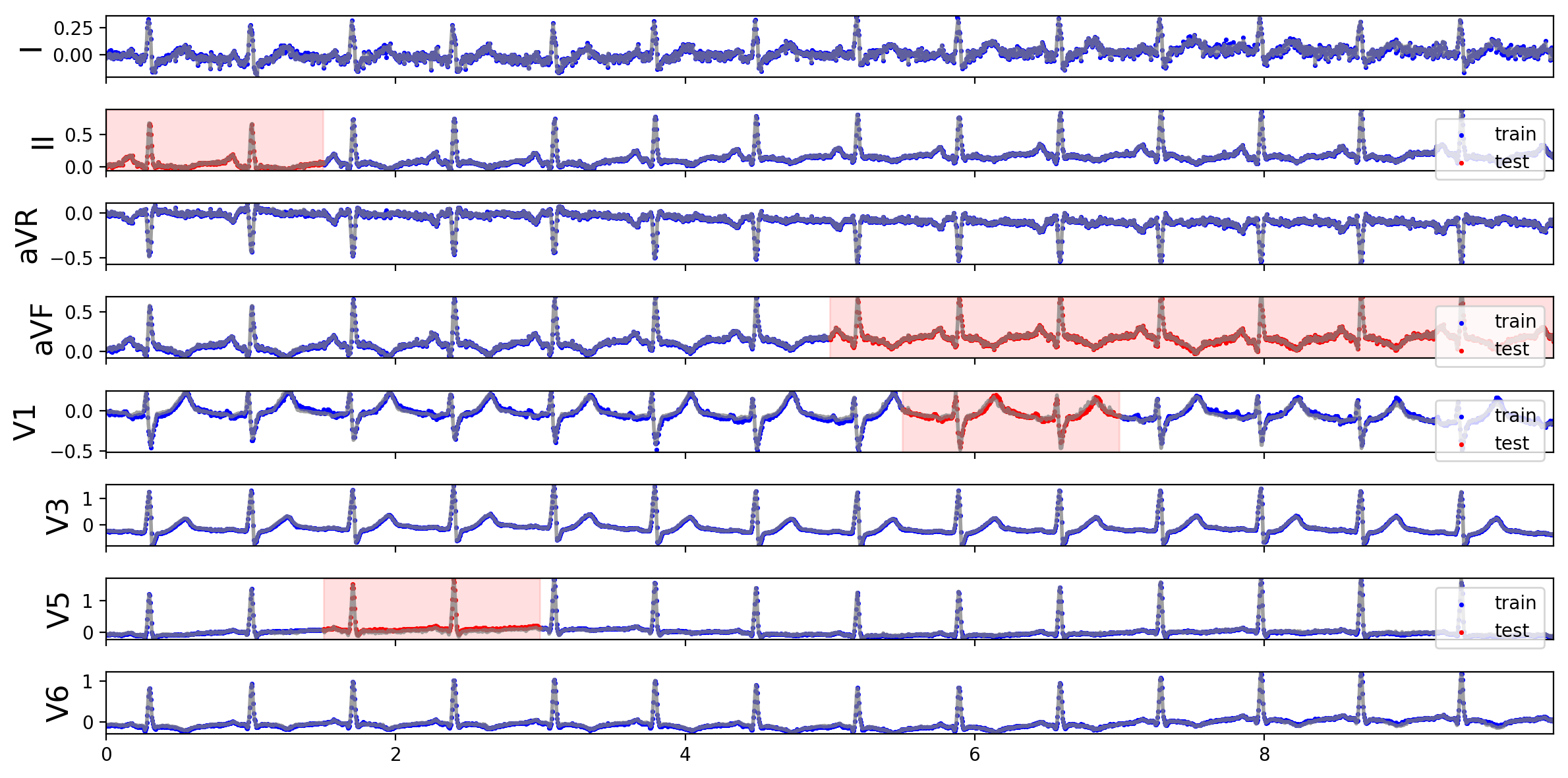}
        \caption{Dipole model}
    \end{subfigure}
    
    \begin{subfigure}[b]{0.9\linewidth}
        \centering
        \includegraphics[width=\textwidth]{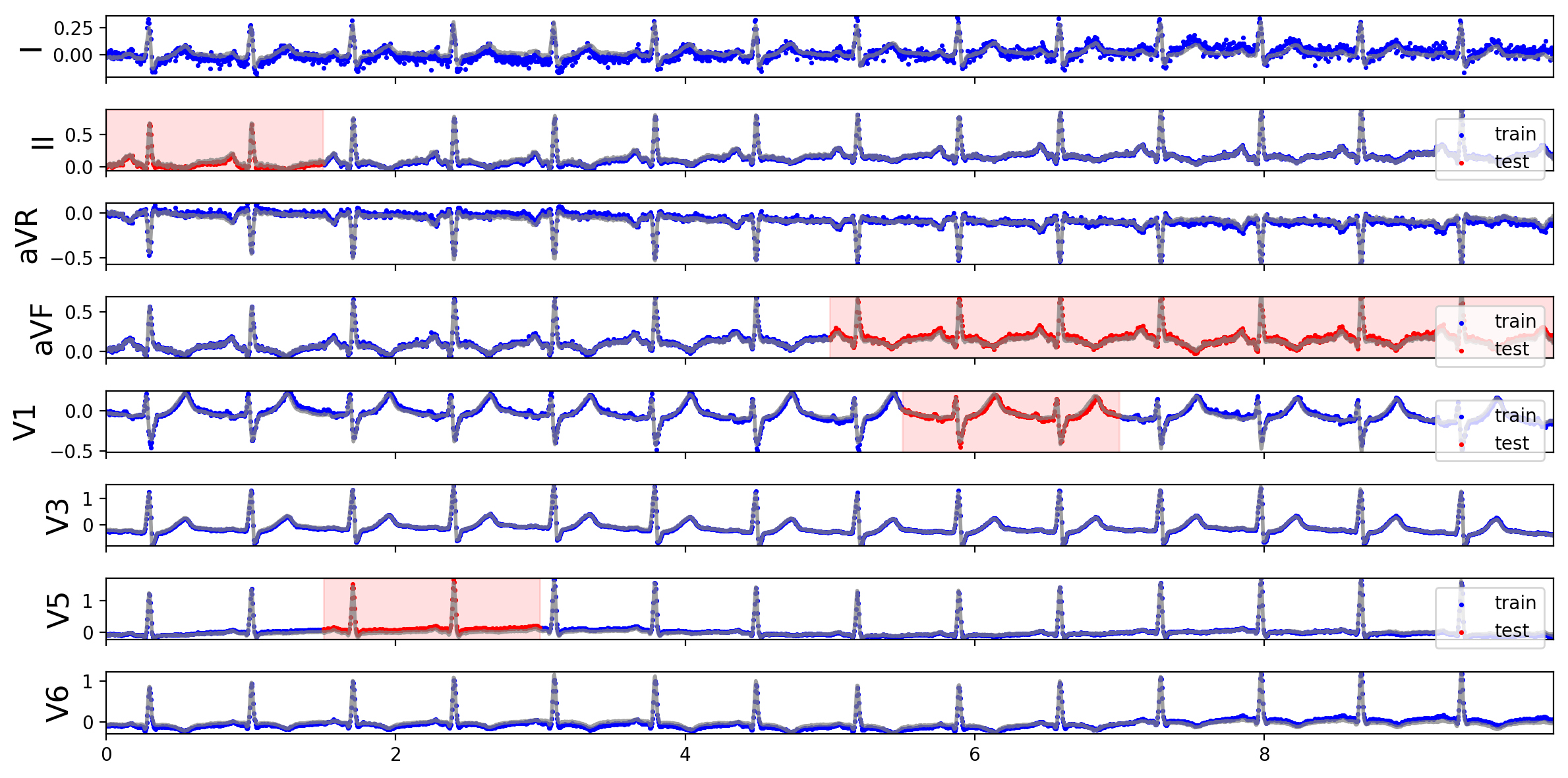}
        \caption{PCA $K=3$}
    \end{subfigure}%
        
    \begin{subfigure}[b]{0.9\linewidth}
        \centering
        \includegraphics[width=\textwidth]{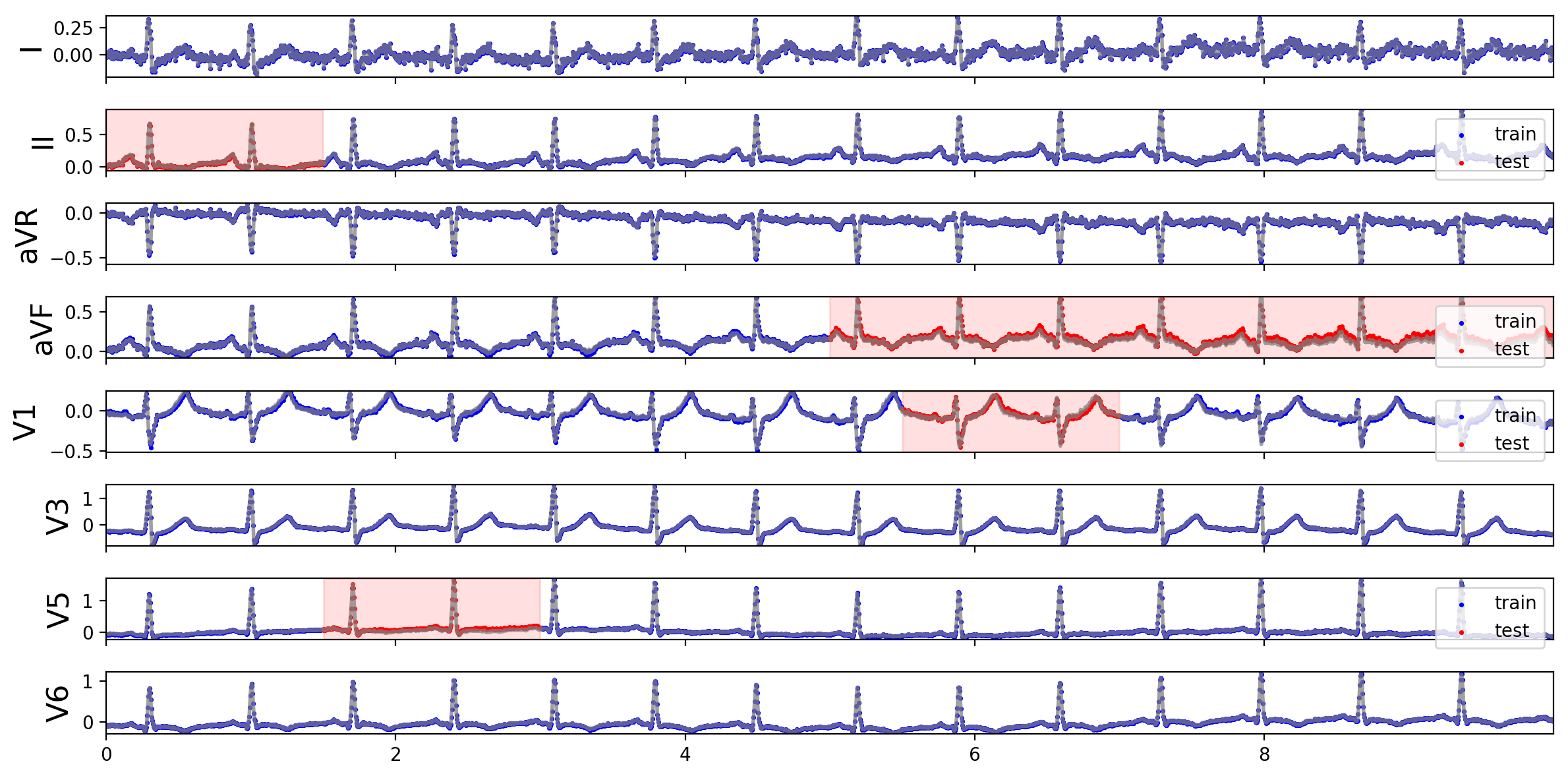}
        \caption{PCA $K=6$}
    \end{subfigure}%
    \caption{
    Held out and missing reconstruction for \texttt{ptb} data (depicting only 8 leads). 
    }
    \label{fig:full-ptb-reconstruction}
\end{figure}

Figure~\ref{fig:model-fit} depicts held-out data, showing the reconstruction performance of the dipole model in leads V1, V5, II and aVF.
Figure~\ref{fig:test-residual-ed} also depicts the residuals for the dipole model and both PCA models.
Figure~\ref{fig:full-ed-reconstruction} displays the full missing data imputation for an emergency department patient. 
Similarly, Figure~\ref{fig:full-ptb-reconstruction} shows the missing data imputation for the \texttt{ptb} example, which has no missing data. 

%---------------------------------%
% Hold-out residual figure        %
%---------------------------------%

%---------------------------------%
% Full imputation figure          %
%---------------------------------%

\clearpage
\newpage
\section{Inference}
\label{app:inference}
We currently fit the model with MAP inference over both the latent $\bz_t$ and global $\br_e$ parameters.  This joint maximization, however, may lead to overfitting --- we would prefer to maximize the following marginal likelihood objective
\begin{align}
\{\hat \br\} 
  &= \argmax_{\br} \ln p(\bx | \br) p(\br) \\
  &= \argmax_{\br} \ln \int p(\bx | \bz, \br) p(\bz) p(\br) d\bz \, .
\end{align}
As an example, this is the objective that PCA essentially fits --- the local weights are integrated out (in the E-step) and the global factors are optimized (in the M-step). 

Our dipole is challenging, however, because the latent factors $\bz$ and global variables $\br$ enter into the likelihood term, $p(\bx | \bz, \br)$ with a complex function, given by Equation~\ref{eq:likelihood}. 

The posterior distribution over $\bz$ given the observations $\bx$ and electrode locations $\br$ does not admit a closed form
\begin{align}
    p(\bz_t | \bx_t , \br) &\propto p(\bx_t | \bz, \br) p(\bz_t) \\
    &= \left( \prod_{\ell=1}^{|L|} p(\bx_{t,\ell} | \bz_t, \br) \right)  p(\bz_t) \\
    &= \left( \prod_{\ell=1}^{|L|} \mathcal{N}(\bx_{t,\ell} | 
    (\bO_{\ell} \cdot \bg(\bz_t, \br),  \sigma^2 \right) p(\bz_t) 
\end{align}
where we define $\bg$ to be the $|E|$-length electrode observations 
\begin{align}
    \bg(\bz_t, \br) &= \left(g(\bs_t, \bp_t, \br_1, \kappa), \dots, g(\bs_t, \bp_t, \br_{|E|}, \kappa) \right)^\intercal \, , 
\end{align}
recalling that the latent state $\bz_t \triangleq (\bs_t, \bp_t)$ is divided into a location and moment component.
Because $g(\bs_t, \bp_t, \br_{e}, \kappa)$ is nonlinear in $\bs_t$, the posterior distribution over $\bs_t$ becomes intractable.  
We do note that $g(\cdot)$ is linear in $\bp_t$ when you fix $\bs_t$, $\br_e$, and $\kappa$.
If we place a Gaussian prior over $\bp_t$, this conditional posterior should be tractable.
We plan to use this tractable conditional structure in future work to scale inference in this model.

\end{document}